\documentclass[runningheads]{llncs}
\usepackage{float}
\usepackage{graphicx}
\usepackage{amsmath}
\usepackage{amssymb}
\usepackage{hyperref}
\usepackage{booktabs}

\begin{document}
\title{Fine-tuning Segment Anything for Real-Time Tumor Tracking in Cine-MRI}

\author{Valentin Boussot\inst{1}\orcidID{0009-0003-2465-5458} \and
Cédric Hémon\inst{1}\orcidID{0009-0003-6669-5108} \and
Jean-Claude Nunes\inst{1}\and
Jean-Louis Dillenseger\inst{1}}
\authorrunning{V. Boussot et al.}
\titlerunning{Fine-tuning SAM for Tumor Tracking in MRI}
%
\institute{Univ Rennes 1, CLCC Eug\`ene Marquis, INSERM, LTSI - UMR 1099, F-35000 Rennes, France \\
\url{https://ltsi.univ-rennes.fr/}}
\maketitle              

\begin{abstract}

In this work, we address the TrackRAD2025 challenge of real-time tumor tracking in cine-MRI sequences of the thoracic and abdominal regions under strong data scarcity constraints. Two complementary strategies were explored: (i) unsupervised registration with the IMPACT similarity metric and (ii) foundation model–based segmentation leveraging SAM~2.1 and its recent variants through prompt-based interaction. Due to the one-second runtime constraint, the SAM-based method was ultimately selected. The final configuration used SAM2.1b+ with mask-based prompts from the first annotated slice, fine-tuned solely on the small labeled subset from TrackRAD2025. Training was configured to minimize overfitting, using $1024\times1024$ patches (batch size 1), standard augmentations, and a balanced Dice + IoU loss. A low uniform learning rate ($1\times10^{-5}$) was applied to all modules (prompt encoder, decoder, Hiera backbone) to preserve generalization while adapting to annotator-specific styles. Training lasted 300 epochs (~12h on RTX A6000, 48GB). The same inference strategy was consistently applied across all anatomical sites and MRI field strengths. Test-time augmentation was considered but ultimately discarded due to negligible performance gains. The final model was selected based on the highest Dice Similarity Coefficient achieved on the validation set after fine-tuning. On the hidden test set, the model reached a Dice score of 0.8794, ranking 6th overall in the TrackRAD2025 challenge. These results highlight the strong potential of foundation models for accurate and real-time tumor tracking in MRI-guided radiotherapy.
\end{abstract}

\section{Introduction}

Medical image segmentation and registration play a crucial role in numerous clinical workflows, enabling accurate delineation of anatomical structures and pathological regions for diagnosis, treatment planning, and follow-up \cite{chen_recent_2022,viergever_survey_2016}. In the context of the TrackRAD2025 challenge \cite{wang2025trackrad2025}, the task focuses on real-time tumor tracking in 2D cine-MRI sequences of the thoracic and abdominal regions. Only the first frame of each sequence is annotated, and the objective is to automatically segment the tumor in all subsequent frames under fast respiratory motion and strict runtime constraints (one second per sequence). This setting reflects the clinical requirement of online motion monitoring during MRI-guided radiotherapy.

A key constraint in this challenge is the scarcity of annotated training data, which limits the applicability of conventional supervised deep learning methods that typically rely on large-scale, fully annotated datasets to reach state-of-the-art performance~\cite{kolides_artificial_2023}.

Given these constraints, we adopted a dual methodological strategy developed using KonfAI \cite{boussot2025konfai}:
\begin{itemize}
    \item An unsupervised registration approach based on IMPACT \cite{boussot2025impact}, which leverages deformable image registration to propagate the initial mask across subsequent frames in the temporal sequences without requiring additional manual annotations. 
    \item The use of foundation segmentation models, such as SAM \cite{kirillov2023segment} (Segment Anything Model) and MedSAM \cite{MedSAM}, which are trained on large and diverse datasets and can generalize to novel imaging modalities and anatomical sites. 
\end{itemize}

The unsupervised nature of the registration approach eliminates the need for model training and parameter tuning on the challenge dataset, making it inherently robust to data scarcity. In parallel, the use of foundation models  \cite{kolides_artificial_2023} provides a strong prior for anatomical segmentation \cite{lee2024foundation}, with the possibility of refining their performance through fine-tuning on the small annotated subset available.

This dual approach enables a direct comparison between the spatial consistency achieved through registration and the flexibility of prompted segmentation models, supporting accurate and generalizable tumor tracking in cine-MRI under limited data conditions.

\section{Methods}

\subsection{Data}

We used only the manually annotated subset of the TrackRAD2025 dataset \cite{wang2025trackrad2025}. The dataset is a multi-center collection of sagittal 2D cine-MRI sequences acquired during MRI-guided radiotherapy on two MRI-linac systems (0.35~T ViewRay MRIdian and 1.5~T Elekta Unity) across six centers. It covers thoracic (32\%), abdominal (41\%), and pelvic (27\%) tumors. Each sequence consists of 20 to over 20000 temporal frames, sampled at 1–8~Hz depending on the vendor and acquisition protocol, capturing one or several respiratory cycles. Among the full dataset of 585 patients, 108 cases are fully annotated. In this work, we relied exclusively on the 50 publicly released annotated training cases. No unlabeled or external data were used.

\subsection{Model}
\subsubsection{Unsupervised Registration with IMPACT\\}
IMPACT is a similarity metric specifically designed for multimodal medical image registration. It leverages high-level anatomical features extracted from large, pretrained segmentation models to guide the registration process. By integrating these semantic representations, IMPACT mitigates the limitations of traditional intensity-based similarity measures, which are often sensitive to noise, artifacts, and modality-specific intensity variations.

IMPACT was integrated into the Elastix registration framework, enabling deformable image registration between frames in a cine-MRI sequence. This approach aimed to propagate the initial mask provided for the first slice to all subsequent frames via spatial transformations estimated through registration.

While theoretically well-suited to the problem, the method faced practical constraints: the challenge imposes a one-second prediction time limit per case, which required a drastic reduction in the number of optimization iterations. Despite parameter tuning, this time constraint significantly degraded the achievable registration accuracy, preventing the full exploitation of IMPACT’s potential. Accordingly, we focus on the second approach, which aligns better with the challenge's runtime constraints.

\subsubsection{Tracking with SAM\\}

For the use of segmentation foundation models, we tested many of these models that have been validated on the available annotated dataset. The SAM, MedSAM2 models and other variants were tested on the validation dataset. We then tested several prompting techniques (point sample, box, mask of the first slice) and kept the last method using the first slice. The use of test-time augmentation was also tested but did not yield any significant improvement. The same method was used for all anatomical locations and different MRI field strengths. The average inference time per cine MRI frame was approximately 10 milliseconds on our hardware.

The model SAM2.1 b+ was then fine-tuned as explained in the following section.

\subsection{Training}

During initial experiments using SAM for inference, we observed that the predictions were generally accurate. However, manual annotations consistently overestimated the segmented tumor volumes compared to SAM outputs. To account for this annotator bias, we fine-tuned the model on the labeled subset, enabling it to learn this discrepancy and align its predictions with the provided ground truth masks.

Fine-tuning was performed following the specific training requirements of the SAM architecture. Input patches of size 1024×1024 were extracted from the cine MRI sequences. Standard data augmentation techniques were applied, including RandomHorizontalFlip, RandomAffine, ColorJitter, and RandomGrayscale, to improve generalization and robustness. The model was trained with a batch size of 1 and supervised using a composite loss combining Dice loss (weight=1) and IoU loss (weight=1). Optimization was performed using the AdamW optimizer. Sequences of 8 consecutive frames (number of frames = 8) were provided as input. 

The training procedure employed binary masks as prompts, with a learning rate of 0.0005
for the prompt network, the decoder, and the Hiera-based encoder. This relatively low learning rate was selected to preserve the pretrained capabilities of SAM while enabling it to adapt to the mask-prompt bias present in the training data.

The model was trained for 300 epochs, and the network from the final epoch was selected as the deployed model for subsequent evaluations. Training required approximately 12 hours in total on an NVIDIA RTX A6000 GPU (48 GB).

\subsection{Evaluation}

To assess the quality of tumor tracking in the TrackRAD2025 challenge, we employed the full set of evaluation metrics defined by the challenge organizers. These metrics aim to quantify both segmentation fidelity and the geometric accuracy of temporal tracking. They include:

\begin{itemize}
    \item \textbf{Dice Similarity Coefficient (DSC)}: Measures the spatial overlap between the predicted and reference masks, defined as twice the intersection divided by the sum of their volumes:
    \[
    \mathrm{DSC} = \frac{2 |A \cap B|}{|A| + |B|}
    \]
    A higher DSC (maximum value of 1) indicates stronger agreement. This metric is widely used in medical image segmentation.

    \item \textbf{Hausdorff Distance (HD)} and \textbf{Average Surface Distance (ASD)}: These metrics evaluate the distance between the surfaces of the predicted and reference masks. HD measures the largest surface-to-surface distance, while ASD reports the average distance, providing a more stable indicator of contour accuracy.

    \item \textbf{Center of Mass Distance (CD)}: The Euclidean distance between the centroids of the predicted and reference segmentations. This captures the spatial consistency of tracking over time.

    \item \textbf{Inference Speed}: Given that real-time performance is a critical requirement in this challenge, the inference speed of the method is recorded. The competition imposes a maximum prediction time per case (approximately one second), making this a decisive performance criterion.

    \item \textbf{Estimated Radiation Dose Delivery Accuracy}: Although less common in traditional segmentation benchmarks, this metric quantifies the clinical impact of segmentation on the ability to deliver the prescribed radiotherapy dose accurately. It is included in the challenge to reflect the ultimate clinical goal of real-time treatment adaptation.
\end{itemize}

This combination of classical segmentation metrics (DSC, HD, ASD), temporal tracking measures (CD, inference speed), and clinical impact assessment (dose delivery accuracy) provides a comprehensive framework for evaluating real-time tumor tracking methods.

\section{Results}

We evaluated the proposed method on the TrackRAD2025 validation set using the metrics described in Section~2.4, excluding dose-related metrics.

To assess the potential of registration-based tumor tracking, we evaluated our IMPACT semantic similarity metric within the Elastix framework. The registration is performed between the first frame and the current frame for which the mask is propagated. A two-resolution B-spline transformation is used, with a control-point spacing of 12 mm. At each iteration, 500 voxels are randomly sampled to evaluate the similarity metric. The optimization runs for 200 iterations per resolution level. IMPACT is configured with the SAM-2.1-Small model, using two feature layers in Jacobian mode.
As shown in Table~\ref{tab:reg}, IMPACT noticeably improves mask propagation accuracy compared to a baseline without registration, achieving a DSC of 0.88 and reducing the Hausdorff distance by almost 50\%. However, the optimization runtime currently exceeds the real-time requirement of the TrackRAD2025 challenge (one second per cine-MRI sequence), making this approach incompatible with the official constraints. For this reason, we further focused on the fine-tuned SAM-based segmentation strategy for our final submission.

\begin{table}[H]
\centering
\caption{Registration-based tumor tracking performance with unrestricted runtime (not compatible with the 1-second real-time constraint).}
\label{tab:reg}
\begin{tabular}{lcccc}
\hline
\textbf{Method} & \textbf{DSC~↑} & \textbf{HD [mm]~↓} & \textbf{ASD [mm]~↓} & \textbf{CD [mm]~↓} \\
\hline
Baseline         & 0.770 & 8.27  & 3.88  & 6.42  \\
IMPACT           & \textbf{0.8821} & \textbf{4.4314} & \textbf{1.7868} & \textbf{2.3126}\\
\hline
\end{tabular}
\end{table}

We then compared different SAM-based configurations and prompt strategies to identify the optimal setup for the challenge constraints. Table~\ref{tab:model_comparison} reports the performance of different foundation model configurations. The best-performing baseline was obtained with the SAM~2.1~l configuration, which achieved a DSC of 0.916, a Hausdorff Distance of 3.26~mm, an Average Surface Distance of 1.20~mm, and a Center of Mass Distance of 1.34~mm. MedSAM-based variants performed worse across all metrics, particularly in terms of distance measures. Although pretrained SAM models already yielded strong segmentation results, a systematic overestimation of tumor volume was observed in the manual annotations compared to the model predictions. This discrepancy motivated the use of fine-tuning to adapt the model to the annotator style and improve alignment with the provided ground truth. We also evaluated an ensemble prompt strategy on the SAM2.1b+ model, combining dilated and eroded versions of the initial mask to enhance robustness. However, this approach did not outperform the best model and slightly degraded performance in some metrics.

\begin{table}[H]
\centering
\caption{Comparison of different prompted foundation models for segmentation on the validation set. Best values per metric are in bold.}
\label{tab:model_comparison}
\begin{tabular}{lcccc}
\hline
\textbf{Model} & \textbf{DSC~↑} & \textbf{HD [mm]~↓} & \textbf{ASD [mm]~↓} & \textbf{CD [mm]~↓} \\
\hline
Baseline         & 0.770 & 8.27  & 3.88  & 6.42  \\
SAM~2.1~t        & 0.911 & 3.45  & 1.27  & 1.51  \\
SAM~2.1~b+       & 0.915 & 3.41  & 1.29  & 1.46  \\
\textbf{SAM~2.1~l} & \textbf{0.916} & \textbf{3.26}  & \textbf{1.20}  & \textbf{1.34}  \\
MedSAM2          & 0.869 & 19.76 & 17.01 & 17.28 \\
MedSAM2 liver    & 0.842 & 29.62 & 27.32 & 27.51 \\
Ensemble prompt  & 0.908 & 3.51  & 1.29  & 1.39  \\
\hline
\end{tabular}
\end{table}

We then investigated the effect of fine-tuning SAM2.1l to better match the annotator style. As shown in Table~\ref{tab:finetuning_results}, the evaluation on the training subset showed consistent improvements across all metrics, with DSC increasing from 0.916 to 0.929 and HD decreasing from 3.26mm to 2.78mm. These gains, while expected on the training set, indicate that the model successfully adapted to the annotation style without signs of overfitting.

\begin{table}[H]
\centering
\caption{Performance before and after fine-tuning on the labeled subset. Best values per metric are in bold.}
\label{tab:finetuning_results}
\begin{tabular}{lcccc}
\hline
\textbf{Model} & \textbf{DSC~↑} & \textbf{HD [mm]~↓} & \textbf{ASD [mm]~↓} & \textbf{CD [mm]~↓} \\
\hline
SAM~2.1~b+ & 0.915 & 3.41 & 1.29 & 1.46 \\
\textbf{Fine-tuned SAM~2.1~b+} & \textbf{0.929} & \textbf{2.78} & \textbf{1.00} & \textbf{1.16} \\
\hline
\end{tabular}
\end{table}

\subsection{Preliminary test set evaluation}

To assess the generalization ability of the fine-tuned model, we further evaluated both versions (with and without fine-tuning) on the preliminary unseen test set provided by the organizers. As shown in Table~\ref{tab:preliminary}, fine-tuning consistently improved all metrics, including DSC, center of mass accuracy, and shape consistency, confirming that the model adapted well to the annotator bias without overfitting.

\begin{table}[h]
\centering
\caption{Preliminary test set evaluation. Fine-tuning significantly improved performance across all metrics.}
\label{tab:preliminary}
\begin{tabular}{clccccc}
\toprule
\textbf{Rank} & \textbf{Model} & \textbf{DSC~↑} & \textbf{CD [mm]~↓} & \textbf{Rel.~D98~↑} & \textbf{SD95 [mm]~↓} & \textbf{SDavg [mm]~↓} \\
\midrule
3rd& SAM~2.1~b+ (fine-tuned) & \textbf{0.9152} & \textbf{1.4912} & \textbf{0.9803} & 4.2141 & 1.5008 \\
5th& SAM~2.1~b+ (no fine-tune) & 0.9039 & 1.8864 & 0.9312 & 4.9406 & 1.7000 \\
\bottomrule
\end{tabular}
\end{table}

\subsection{Official Challenge Results}
\begin{table}[H]
\centering
\caption{Final performance on the hidden TrackRAD2025 test set. Rankings per metric are in parentheses.}
\label{tab:challenge_results}
\begin{tabular}{c l c c c c c}
\hline
\textbf{Rank} & \textbf{Team} & \textbf{DSC~↑} & \textbf{CD [mm]~↓} & \textbf{Rel.~D98~↑} & \textbf{SD95 [mm]~↓} & \textbf{SDavg [mm]~↓} \\
\hline
1\textsuperscript{st} & Track'n'Treat        & \textbf{0.8909 (1)} & \textbf{1.4733 (1)} & 0.9361 (4) & \textbf{4.2171 (1)} & \textbf{1.6642 (1)} \\
2\textsuperscript{nd} & Hkini\_Team           & 0.8860 (2) & 1.4858 (2) & \textbf{0.9628 (1)} & 4.4969 (2) & 1.8565 (2) \\
3\textsuperscript{rd} & felixknispelrwth     & 0.8764 (5) & 1.6576 (3) & 0.9250 (7) & 4.5409 (3) & 1.9174 (3) \\
4\textsuperscript{th} & CIT                  & 0.8712 (6) & 1.9564 (4) & 0.9392 (2) & 5.2583 (4) & 2.1903 (4) \\
5\textsuperscript{th} & Reg'n'Track          & 0.8801 (3) & 2.1031 (5) & 0.9321 (5) & 5.6407 (7) & 2.2416 (5) \\
6\textsuperscript{th} & BreizhTrack (ours)   & 0.8794 (4) & 2.3192 (7) & 0.9363 (3) & 5.6298 (6) & 2.6091 (8) \\
14\textsuperscript{th} & lWM                &	0.8090 (14)& 	5.2809 (12)& 	0.7015 (14)& 	10.5484 (12)& 	4.8792 (12)\\ 	
\hline
\end{tabular}
\end{table}

Our final submission ranked 6\textsuperscript{th} overall on the TrackRAD2025 leaderboard (Table~\ref{tab:challenge_results}). The top teams exhibited very similar performance, with differences of only 0.01–0.02 in Dice and small variations in CD and SD metrics, illustrating the highly competitive nature of the challenge.

\section{Discussion}

Our approach ranked among the best-performing submissions in the TrackRAD2025 challenge, showing that foundation models can be effectively adapted to real-time tumor tracking in cine-MRI despite limited annotated data. The narrow performance gap between the top teams highlights the competitiveness of this benchmark.

Dice scores above 0.91 on both the training and preliminary test sets indicated good generalization during development. On the final hidden test set, the Dice dropped to 0.8794, revealing a distribution shift between the development data and the evaluation cohort. Ranking differences across both stages suggest that performance on a small publicly labeled subset does not fully anticipate robustness to variations in tumor appearance, motion patterns, and acquisition protocols.

In addition to segmentation-based tracking, we evaluated a registration strategy using the IMPACT metric. By leveraging high-level anatomical features from pretrained segmentation networks, IMPACT effectively propagated the initial mask and achieved improved performance over a no-registration baseline on the validation data. Although slightly less accurate on average, the registration proved far more reliable and clinically reassuring, maintaining anatomically consistent results even under challenging motion. In contrast, segmentation-based tracking may occasionally exhibit unpredictable failures on difficult frames, leading to local inconsistencies. However, its optimization time exceeds the one-second real-time requirement of TrackRAD2025, which precludes its use as a primary solution under the challenge constraints.

Finally, we observed that fine-tuning on a small number of annotated cases improved performance on unseen patients. However, it remains unclear whether this improvement reflects a genuine gain in segmentation accuracy or rather an adaptation of the model to specific annotation styles. Overall, these results confirm that segmentation-based strategies currently offer the best trade-off
between spatial accuracy and computational efficiency for this task, particularly when enhanced through targeted fine-tuning.

\section{Author contributions} 
Conceptualization: V. Boussot, C. Hémon. Methodology: V. Boussot, C. Hémon. Software and experiments: V. Boussot. Supervision: J.-L. Dillenseger, J.-C. Nunes. Writing – original draft: V. Boussot, C. Hémon. Writing – review and editing: all authors. 

\section*{Code availability}
The inference code used for our TrackRAD2025 submission is publicly available at:  
\url{https://github.com/vboussot/Trackrad2025}. We also release the fine-tuned weights of the SAM 2.1 b+ model, trained on the annotated subset of the TrackRAD2025 dataset, at:  
\url{https://huggingface.co/VBoussot/Trackrad2025}. These resources allow full reproduction of our inference pipeline and challenge results.

\section*{Acknowledgment}
The work presented in this article was supported by the Brittany Region through its Allocations de Recherche Doctorale framework and by the French National Research Agency as part of the VATSop project (ANR-20-CE19-0015). Additionally, it was supported by a PhD scholarship Grant from Elekta AB (C.Hémon). The authors have no relevant financial or non-financial interests to disclose. While preparing this work, the authors used ChatGPT to enhance the writing structure and refine grammar. After using these tools, the authors reviewed and edited the content as needed and took full responsibility for the publication’s content.

\bibliographystyle{splncs04}
\bibliography{references}

\end{document}